\documentclass{article}
\usepackage[dblblindworkshop, final]{neurips_2026}
\workshoptitle{Foundations of Agentic Systems Theory}

\usepackage[utf8]{inputenc}
\usepackage[T1]{fontenc}
\usepackage[pdfusetitle]{hyperref}
\usepackage{url}
\usepackage{booktabs}
\usepackage{amsfonts}
\usepackage{amsmath}
\usepackage{amssymb}
\usepackage{stmaryrd}
\usepackage{graphicx}
\usepackage{microtype}
\usepackage{array}
\usepackage{tabularx}

\newcommand{\op}[1]{\textsc{#1}}
\newcommand{\arm}[1]{\textsc{#1}}
\newcommand{\code}[1]{\texttt{#1}}

\title{Nothing Is Harder Than Nothing,\\%
Until You Write It Down:\\%
Absence Conventions in Stateful Tool Use}

\author{%
  Marek \v{S}uppa\\
  Comenius University in Bratislava\\
  Cisco Systems\\
  \texttt{marek@mareksuppa.com}
  \And
  Jaroslav Kop\v{c}an\\
  Kempelen Institute of Intelligent Technologies\\
  \texttt{jaroslav.kopcan@kinit.sk}
}

\hypersetup{
  hidelinks,
  pdfauthor={Marek \v{S}uppa, Jaroslav Kop\v{c}an},
  pdfsubject={Foundations of Agentic Systems Theory (FAST) workshop at NeurIPS 2026},
  pdfkeywords={LeaveItOutBench, large language models, LLM agents, function calling,
    stateful tool use, JSON update semantics, omitted fields, null values, empty values,
    tool documentation}
}

\begin{document}
\maketitle

\begin{abstract}
When a tool updates a record, leaving a field out can keep it unchanged or
reset it to a default. LeaveItOutBench tests how large language models (LLMs)
handle omitted fields, empty or falsy values, explicit \code{null} and values
that match the default in stateful function calling. Each task pair changes
one update rule for the same request, schema and target state. Models write
one JSON call as text. We score the full state after the call and distinguish
stored values from inherited defaults.

When the tool description leaves the rule unstated, both tasks use the same
prompt. Across five pre-registered models and 120 held-out task families,
exact accuracy was $76.2\%$ under conventional rules and $49.5\%$ under
reversed rules, a gap of $26.7$ points. On the 32 omission families, accuracy
was $76.9\%$ when omission kept the field unchanged and $3.8\%$ when it reset
the field to a default. Defining the operations and adding a rule table
raised accuracy under reversed rules by $39.8$ percentage points without
worked examples and $43.7$ with them. Three later models from two vendors
also showed a gap under the short description. In a follow-up on Qwen3-14B
alone, two interfaces with explicit operations gained $5.8$ and $6.7$ points
over full documentation. Both fell short of the pre-registered eight-point
threshold.
\end{abstract}

\section{Introduction}

An LLM agent that edits stored records through a tool call (function
calling) has to decide what an omitted field, an empty value or a \code{null}
means, and the protocols it learned from answer differently in their own
specifications. RFC~7396 removes a member whose patch value is \code{null},
so it cannot store a literal \code{null}~\citep{rfc7396}, while RFC~6902 names
\code{remove} as a separate operation~\citep{rfc6902}. GraphQL distinguishes an
omitted input from a supplied \code{null}~\citep{graphql}. A JSON Schema
\code{default} is only an annotation and stores nothing~\citep{jsonschema},
and under Protocol Buffers' implicit presence~\citep{protobuf} a field written
with its default value cannot be told apart from an unset one. Each rule is
valid within its own protocol, so an agent has to follow whichever rule the
tool it is using documents.

\paragraph{An example.} A workspace default supplies a signature line for a
contact that has none. The user asks for a blank signature. Two deployments of
the same update tool differ in how they interpret an empty value. The call
\code{\{"field":"signature\_line","value":""\}} stores the empty string in
the first deployment and clears the field in the second. The second requires
an extra \code{"literal":true} to store the empty string. We call tasks whose
contracts differ in one rule a \emph{minimal pair}. Storing the current default
and resetting the field to inherit it display the same value, so we score the
full stored state.

\paragraph{Contributions.} We measure how exact state accuracy changes when
an unstated update rule reverses a familiar convention. We then test how much
documenting the operations and their rules helps, how much worked examples
add and whether an explicit operation field improves on full documentation.
We pre-registered these tests. Our deterministic compiler turns declarative
update specifications into matched schemas, prompts, reference calls and
exact scorers.

\section{Update algebra and convention inversion}
\label{sec:algebra}

A field can be absent, hold a value of its own or inherit a default. A stored
value and an inherited default can look the same, so we track how each field
gets its value. We represent a field's \emph{latent state} as
$S_V = \operatorname{Absent} \mid \operatorname{Override}(v) \mid
\operatorname{Inherited}(d, r)$, where $v$ is the field's own value and $d$
is its effective default at policy revision $r$.
An update applies one of four operations,
$\Delta(V) = \operatorname{Keep} \mid \operatorname{Set}(v) \mid
\operatorname{Clear} \mid \operatorname{Reset}$, which leave the state
unchanged or produce $\operatorname{Override}(v)$, $\operatorname{Absent}$ and
$\operatorname{Inherited}(d_{\mathrm{cur}}, r_{\mathrm{cur}})$, respectively.
Setting \code{""}, \code{false}, $0$ or \code{null} is an ordinary \op{Set},
never an alias for \op{Clear}. A tool call assigns an operation to every
editable field, including the fields it does not mention, so \op{Keep} is
tested inside a call the model must make.

A \emph{contract} is a set of ten \emph{rules}, each assigning one operation
to an input form of an update entry: omitted, named without a value,
\code{null}, the empty or falsy value, the current default, any other value,
three explicit \code{absence\_mode} flags and \code{literal:true}
(Table~\ref{tab:rows}). Four are \emph{collision rules} (omitted, \code{null},
empty and default): each of their two readings is the dominant convention of a
real protocol family. For example, a bare \code{null} is \op{Clear} under
RFC~7396 and \op{Set}(\code{null}) under GraphQL; Table~\ref{tab:rows} lists
all eight readings with sources. A pair targets one
collision rule: the \emph{aligned} contract reads it as the dominant convention
does, the \emph{counter-conventional} contract the other way, and
every other rule is the same. Five \emph{explicit channels} (a field named
without a value, the three \code{absence\_mode} flags and \code{literal:true})
mean the same under both contracts, so both can express every operation.

The compiler emits 180 task \emph{families}, 60 for development and 120 held
out, split by sentence template and record so that no held-out prompt reuses
a development template (Appendix~\ref{app:contracts}). Each family yields an
aligned and a counter-conventional instance, a minimal pair, that share intent,
initial state, defaults, target state and per-interface schema. The
\emph{convention-following call} is the call a model writes if it follows the
dominant convention; under the counter-conventional contract it reaches a
\emph{displaced state}. The compiler keeps a family only if the aligned
reference call is the convention-following call and the displaced state
differs from the target in its latent state. Requests never contain
\code{keep}, \code{set}, \code{clear} or \code{reset}. All
$1800$ reference calls reach their target, all $900$ counter-conventional
convention-following calls reach their displaced state, and $12{,}167$ audit
checks pass (Appendix~\ref{app:contracts}).

\section{Five interfaces and exact scoring}
\label{sec:interfaces}

We compare three descriptions of the same update tool and two schemas that
require an explicit operation. The model can write one JSON call as text or
abstain. The task ends after that response. The first three interfaces share
a schema that accepts \code{value}, \code{absence\_mode} and \code{literal}.
\arm{ValueOnly}, the \emph{terse description}, leaves the tested rule
unstated. The prompts are identical within each pair, and the rule used to
execute and score the call differs. \arm{Value+Prose} adds the four operation
definitions, the ten rules of the active contract as a truth table, and one
worked example per legal transition on a demonstration field.
\arm{Value+Prose(no examples)} drops the examples.
\arm{FlatDelta} requires an \code{op} field with the four operation names, so
\code{\{"op":"clear","value":x\}} can still be written and is rejected.
\arm{TypedDelta} is a tagged union in which only \code{set} carries a value,
so that entry cannot be written at all (schemas and description lengths in
Appendix~\ref{app:interfaces}).

Scoring depends only on the raw response and the task. We
extract the call from a fenced block, the whole response or the first balanced
brace span, in that order, with no repair, coercion, retry or LLM judge. The
call runs once under the active contract, which also decides what happens to
every field the call does not mention. An output is correct if and only if the
latent state of every field, untouched fields included, equals the target.
Parse failures, malformed wrappers, abstentions and invalid calls count as
incorrect. The \emph{convention-inversion penalty} (CIP) is aligned minus
counter-conventional accuracy on the same families, with bootstrap intervals
over families (Appendix~\ref{app:protocol}).

\begin{figure}[t]
\centering
\includegraphics[width=\textwidth]{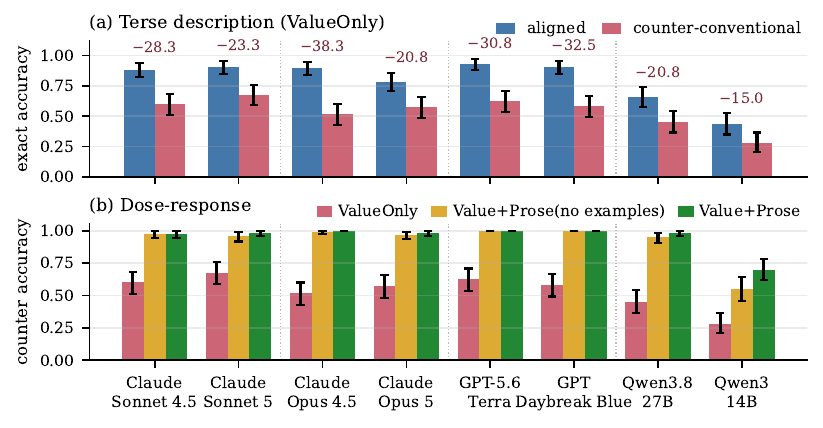}
\caption{Held-out accuracy of all eight models. Dotted lines separate four
pairs: each pre-registered commercial model with its newer sibling, then the
two local models. (a) Under the terse description, inverting one rule costs
every model the annotated points. (b) Counter-conventional accuracy as
documentation is added. Operation definitions and a truth table improve
accuracy on every commercial model, with or without examples. Error bars:
95\% bootstrap intervals over 120 families.}
\label{fig:main}
\end{figure}

\begin{table}[t]
\centering
\caption{Pre-registered tests on held-out data, pooled over the five
pre-registered models. Thresholds and the Holm family were fixed before any
held-out prompt was sent; Q3 has no threshold.}
\label{tab:gates}
\scriptsize
\begin{tabular}{llrrrl}
\toprule
& Contrast & $\Delta$ pp & 95\% CI & $p_{\mathrm{Holm}}$ & Verdict \\
\midrule
Q1 & ValueOnly aligned vs counter & 26.7 & [20.7, 33.0] & 4.44e-39 & pass \\
Q2 & Value+Prose vs ValueOnly (counter) & 43.7 & [38.0, 49.3] & 7.15e-77 & pass \\
Q3 & Value+Prose(no ex.) vs ValueOnly (counter) & 39.8 & [34.2, 45.5] & 1.26e-68 & reported \\
Q4 & ValueOnly aligned vs counter, omission only & 73.1 & [63.1, 81.9] & 1.2e-35 & pass \\
\bottomrule
\end{tabular}

\end{table}

\section{Results}
\label{sec:results}

With identical prompts, reversing an unstated rule lowers accuracy most for
omitted fields. Documenting the operations and their rules recovers most of
the lost accuracy, including on newer models. The confirmatory study is 120
held-out families $\times$ 2 conditions $\times$ 3 interfaces $\times$ 5
models $= 3600$ outputs. We pre-registered it after a development pilot
stopped, because \arm{Value+Prose} left the structural interfaces no room to
improve, and a development-only run of \arm{ValueOnly} showed a large penalty
(Appendix~\ref{app:protocol}).

\paragraph{Accuracy under unstated rules.} Pooled over the five
pre-registered models, \arm{ValueOnly} scores $76.2\%$ on aligned and
$49.5\%$ on counter-conventional instances, a penalty of $26.7$ points
$[20.7, 33.0]$ (Table~\ref{tab:gates}). The largest penalty, $38.3$ points
($90.0\%\rightarrow51.7\%$), belongs to \texttt{claude-opus-4-5}, with $28.3$
and $30.8$ points on the other commercial models (Figure~\ref{fig:main}a), so
the penalty is not limited to small models. Of the counter-conventional
\arm{ValueOnly} errors, $45.5\%$ ($138/303$, 95\% CI $[39.8, 51.3]$) reach
exactly the displaced state.

\paragraph{Rules help without worked examples.} The pilot's \arm{Value+Prose}
also had worked examples, which could have supplied the answer. On
counter-conventional instances,
\arm{Value+Prose(no examples)} recovers $39.8$ points $[34.2, 45.5]$ over
\arm{ValueOnly}, against $43.7$ points $[38.0, 49.3]$ for \arm{Value+Prose},
a gap of $3.8$ points inside the pre-registered $5$-point band
(Figure~\ref{fig:main}b). Both \arm{Value+Prose} interfaces stay non-inferior on
aligned instances ($97.5\%$ and $97.3\%$ against $76.2\%$).

\paragraph{Newer models show the same failure.} A third pre-registration
added the newer \code{claude-opus-5} and \code{claude-sonnet-5} with the same
two thresholds ($1440$ outputs, same families).
Their pooled penalty is $22.1$ points $[14.6, 30.0]$, and the truth table
recovers $35.8$ points $[28.7, 43.3]$. As pre-registered, we report that the
penalty is smaller than for their predecessors: $38.3\rightarrow20.8$ points
from \code{claude-opus-4-5} to \code{claude-opus-5} and $28.3\rightarrow23.3$
for the Sonnet pair. For the Opus pair, however, counter-conventional accuracy
rises only $5.8$ points while aligned accuracy falls $11.7$ points and invalid
calls under the terse description reach $14.2\%$: the gap narrows mainly
because the newer model does worse on the terse task overall. A fourth
pre-registration added the second vendor's newest model
(\code{gpt-5.6-terra}$\rightarrow$\code{gpt-daybreak-blue-latest}, $720$
outputs): the penalty is $32.5$ points $[22.5, 42.5]$ ($30.8\rightarrow32.5$
across the pair), and the truth table recovers
$41.7$ points $[33.3, 50.8]$. The penalty is present in the newest published
model of both API vendors, and the smaller Anthropic penalty is specific to
those two models.

\paragraph{The penalty is concentrated in omitted fields.} It is
$73.1$ points $[63.1, 81.9]$ on \code{omission}, where counter-conventional
accuracy is $3.8\%$, against $5.7$--$13.1$ points on \code{null},
\code{empty} and \code{default} (Appendix~\ref{app:tables}). When omission
means keep, the reference call can leave unchanged fields out. When omission
means reset, the call must name every field it needs to keep in all five
interfaces. The comparison changes both the rule and the number of fields a
correct call must name. Under \arm{ValueOnly} an explicit \op{Keep} appears in
$26$ of $400$ such slots, with $306$ omitted; with the truth table,
$327$ and $60$; without examples, $331$ and $61$.

\paragraph{Structural interfaces stay below the bar.} Pre-registered decision
rule 7 (Appendix~\ref{app:protocol}) required us to run \arm{FlatDelta} and
\arm{TypedDelta}, at the original $8$-point threshold, on any model whose
\arm{Value+Prose} counter-conventional accuracy fell below $90\%$. Only the
small local model qualified, at $70.0\%$. Neither interface reaches the
threshold: they gain $+5.8$ points $[2.5, 10.0]$ and $+6.7$ points
$[1.7, 11.7]$ over \arm{Value+Prose} ($480$ outputs). Both intervals exclude
zero. A general benefit from explicit operations is not supported.
\arm{TypedDelta} exceeds \arm{FlatDelta} by $0.8$ points
$[-1.7, 4.2]$, short of the $3$ points a type-level claim needed.

\section{Related work}
\label{sec:related}

Executable tool-use benchmarks such as API-Bank~\citep{apibank},
ToolLLM~\citep{toolllm}, Gorilla and BFCL~\citep{gorilla,bfcl},
AgentBench~\citep{agentbench}, ToolSandbox~\citep{toolsandbox},
$\tau$-bench~\citep{taubench} and Thinkingbox~\citep{thinkingbox} cover state,
multi-turn interaction, domain policies and repeated trials. We isolate one
update rule and distinguish storing a default from inheriting it.
IFEval~\citep{ifeval} tests verifiable instructions. IFEval-FC~\citep{ifevalfc}
extends this idea to tool parameter descriptions. The same JSON can mean
different things across services. When the rule is unstated, models have to
infer how the service interprets the call. Stating a rule that reverses a
familiar convention creates a knowledge
conflict~\citep{chameleon,clasheval}. This is a counterfactual
task~\citep{counterfactual} whose sensitivity to frequency \citet{embers}
predict. Work on interface alignment~\citep{attribution}, output
formats~\citep{speakfreely} and constrained decoding~\citep{outlines} studies
how interfaces affect model outputs. We score what an accepted call does to
the record.

\section{Limitations and conclusion}
\label{sec:limitations}

We test single-turn JSON calls in text under a fixed harness preamble
(Appendix~\ref{app:protocol}). The frequency and cost of these mismatches in
deployed tools remain open. The schema comparison covers only the $14$B local
model, which was thirteen months old and the newest of its size available to
us. When the tool left a rule unstated, models often fell back to the familiar
convention. Stating the operations and their rules improved exact state
accuracy on every commercial model we tested.

\paragraph{Use of AI tools.} We used a large language model for grammatical
and stylistic polishing of the text.

\section*{Acknowledgements}
This work was funded by the Slovak Research and Development Agency under the Contract no. APVV-22-0414, and by the European Union NextGenerationEU through the Recovery and Resilience Plan for Slovakia under the project No. 09I01-03-V04-00007.

\bibliographystyle{plainnat}
\bibliography{references}

\appendix

\section{Contracts: the ten rules and their readings}
\label{app:contracts}

An update entry is a JSON object
\code{\{"field": f, "value"?: v, "absence\_mode"?: m, "literal"?: b\}} with
$m \in \{\code{keep}, \code{clear}, \code{reset}\}$. A contract is a total
function from the ten input forms of Table~\ref{tab:rows} to the four
operations of Section~\ref{sec:algebra}. The aligned contract of a family reads
every rule the way the dominant convention does. The counter-conventional
contract changes exactly the designated collision rule and nothing else. In
every documented interface the two instances of a pair differ in that rule's
sentence, and in \arm{Value+Prose} also in the worked examples it affects
(Appendix~\ref{app:pair}). The reference call differs wherever the rule forces
it: always in the three overloaded interfaces, and in the structural
interfaces only for \code{omission} families.
Falsy values are \code{""} for strings, \code{[]} for lists, \code{\{\}} for
objects, \code{false} for booleans and $0$ for integers. Enumerations have
none, so the \code{empty} family never uses an enumeration field, and the
\code{null} family uses only nullable strings. Because the explicit channels
mean the same under both contracts, every operation can be expressed under
both. When the compiler chooses each instance's reference call, it uses the
designated rule where it can and an explicit channel only when that rule
cannot express the required operation.

\begin{table}[h]
\centering
\caption{The ten rules. Each collision rule has two readings; the other six
mean the same under both contracts. The Aligned and Counter columns give the
forward direction; reverse-direction \code{null}, \code{empty} and
\code{default} families swap the two readings. Sources are the specifications
each reading comes from.}
\label{tab:rows}
\scriptsize
\begin{tabularx}{\textwidth}{l l l l X}
\toprule
Rule & Input form & Aligned & Counter & Source (aligned / counter) \\
\midrule
\code{omitted} & field absent from \code{updates} & \op{Keep} & \op{Reset} &
  RFC~7396 partial patch / full-replacement update \\
\code{bare\_null} & \code{\{"field":f,"value":null\}} & \op{Clear} & \op{Set}(\code{null}) &
  RFC~7396 / GraphQL, protobuf explicit presence \\
\code{bare\_empty} & \code{value} is the falsy inhabitant & \op{Set}(v) & \op{Clear} &
  explicit field presence / protobuf implicit presence \\
\code{bare\_default} & \code{value} equals the current default & \op{Set}(v) & \op{Reset} &
  JSON Schema \code{default} annotation / protobuf implicit presence \\
\midrule
\code{bare\_no\_value} & \code{\{"field":f\}} & \op{Keep} & \op{Keep} & explicit channel \\
\code{bare\_other} & any other permitted value & \op{Set}(v) & \op{Set}(v) & none \\
\code{mode\_keep} & \code{"absence\_mode":"keep"} & \op{Keep} & \op{Keep} & explicit channel \\
\code{mode\_clear} & \code{"absence\_mode":"clear"} & \op{Clear} & \op{Clear} & explicit channel \\
\code{mode\_reset} & \code{"absence\_mode":"reset"} & \op{Reset} & \op{Reset} & explicit channel \\
\code{literal} & \code{"literal":true} with a \code{value} & \op{Set}(v) & \op{Set}(v) & explicit channel \\
\bottomrule
\end{tabularx}
\end{table}

The \code{omission} family has only a forward direction (aligned \op{Keep},
counter \op{Reset}). The \code{null}, \code{empty} and \code{default}
families are compiled in both directions, so all six target operations at the
collision field (\op{Keep}, \op{Clear}, \op{Set}(\code{null}),
\op{Set}(empty), \op{Set}(default) and \op{Reset}) occur. The held-out split
has 32 \code{omission} families, 14 forward and 14 reverse for \code{null} and
\code{empty}, and 16 forward and 16 reverse for \code{default}. The
development split has half of each.

\paragraph{Compile-time rejection.} A candidate instance is emitted only if
every one of the following holds.
\begin{itemize}\setlength{\itemsep}{0pt}
\item The two contracts differ in exactly the designated rule, and each can
  still express all four operations.
\item Every interface's reference call executes and reaches the target exactly.
\item The target differs from the initial state.
\item The convention-following call agrees with the reference call iff the
  condition is aligned. Under the counter contract the displaced state differs
  from the target, and the convention-following call executes to that
  displaced state in every interface where it is valid.
\item Documented interfaces contain all four operation words, the first 48
  characters of each of the ten rule sentences and, when examples are enabled,
  every example on the demonstration field, with no record field name inside
  the examples block.
\item Editable and read-only field sets are disjoint, and every default is
  non-null and distinct from the falsy value.
\end{itemize}
Request text is assembled from clause
templates that must contain marker phrases for each operation and may not
contain \code{keep}, \code{set}, \code{clear}, \code{reset}, \code{override} or
\code{inherit}. We split development and held-out data by the 12 lexical
frames, that is, sentence templates (F01--F04 development, F05--F12 held out),
and by the 48 record entities (4 development and 8 held-out records in each of
the four domains).

\section{A worked held-out minimal pair}
\label{app:pair}

Family \code{T000-omission-fwd-contacts}, in the workspace contact directory
domain with sentence template F05. Both instances share the system prompt, the
state block, the schema and the request below. Their omission rules differ.
The prompts are identical under \arm{ValueOnly}. When the description states
the rules, its omission row and the affected examples differ. The system
prompt asks for exactly one JSON object, either an
\code{update\_record} call or \code{\{"abstain": true, "reason": ...\}}, states
that no tool result and no second turn will follow, and says ``Follow the tool
documentation exactly.'' To save space, the state block below leaves out the
section markers, the service line, the two read-only fields and the field
labels, and puts each type on its field's line. The tool descriptions are
verbatim apart from line wrapping.

{\footnotesize\begin{verbatim}
RECORD: DP-5202 (the directory profile for Iveta Sujanska)
EDITABLE FIELDS:
  signature_line       type: string
    stored: absent (the record stores nothing for this field)
    current workspace directory policy default: "Sent from the studio workspace"
      (revision dir-2031.4)
  reply_to_address     type: string or null
    stored: the record's own pinned value "ops@studio.example"
    current workspace directory policy default: "desk@studio.example"
      (revision dir-2031.4)
  label_tags           type: array of strings
    stored: inherited from workspace directory policy revision dir-2030.9,
      effective value ["directory"]
    current workspace directory policy default: ["directory", "studio"]
      (revision dir-2031.4)
  share_photo          type: boolean
    stored: inherited from workspace directory policy revision dir-2031.4,
      effective value true
    current workspace directory policy default: true (revision dir-2031.4)

USER REQUEST
Apply the following change to the directory profile for Iveta Sujanska. Write
"press@studio.example" into the reply-to address. The signature line is to remain
unchanged, exactly as it is stored today. Leave every other field of the record
exactly as it is.
\end{verbatim}}

\noindent The \arm{ValueOnly} description, in full. The prompt does not
include the service field reference it mentions.

{\footnotesize\begin{verbatim}
update_record(record_id, updates)

Applies a list of field update entries to one record of the workspace contact
directory. Each entry names exactly one editable field of that record.

Entries may supply a `value`, and may use the optional per-entry controls
`absence_mode` and `literal`. Consult the service field reference for how
particular payloads are handled.
\end{verbatim}}

\noindent The \arm{Value+Prose} description of the \emph{counter-conventional}
instance keeps the first paragraph above, replaces the second with the
following, and then gives the JSON schema. The aligned instance's rule 1
instead reads ``An editable field that does not appear in \code{updates} is
left unchanged (keep).'' and its \code{keep} example is
\code{\{"field": "demo\_note", "absence\_mode": "keep"\}}.

{\footnotesize\begin{verbatim}
OPERATIONS
This service records four operations on a field:
  keep  leaves the field exactly as it is, including whether it holds the record's
        own pinned value, inherits from the workspace directory policy default, or
        is absent.
  set  stores the given value as the record's own pinned value. A pinned value does
        not change when the workspace directory policy default is later revised.
  clear  makes the field absent: the record then stores nothing at all for it.
        This is not the same as storing an empty value.
  reset  makes the field inherit from the workspace directory policy default at
        the current policy revision, so it follows later revisions of that default.

PAYLOAD INTERPRETATION RULES FOR THIS DEPLOYMENT
  1. An editable field that does not appear in `updates` is returned to inheritance
     from the workspace directory policy default (reset); this service applies a
     full-replacement update, not a partial patch.
  2. An entry that names a field but supplies no `value` at all leaves the field
     unchanged (keep).
  3. An entry whose `value` is null removes the field, leaving it absent (clear).
  4. An entry whose `value` is the empty or falsy inhabitant of the field's type
     stores that value as the record's own pinned value (set).
  5. An entry whose `value` equals the field's current workspace directory policy
     default stores that value as the record's own pinned value (set), which stays
     put when the default is later revised.
  6. An entry whose `value` is any other permitted value of the field's type stores
     that value as the record's own pinned value (set).
  7. An entry with `absence_mode` "keep" leaves the field unchanged (keep).
  8. An entry with `absence_mode` "clear" removes the field, leaving it absent (clear).
  9. An entry with `absence_mode` "reset" returns the field to inheritance from the
     workspace directory policy default (reset).
  10. An entry with `literal` true stores its `value` exactly as written as the
     record's own pinned value (set), whatever that value is, including null, an
     empty value, or the current default.

Every entry of `updates` is interpreted by the rules above and by nothing else.

EXAMPLES (`demo_note` is a string-or-null field used for illustration only; it is
not a field of the record above)
  keep: {"field": "demo_note"}
  set: {"field": "demo_note", "value": "new text"}
  set (empty value): {"field": "demo_note", "value": ""}
  set (explicit null): {"field": "demo_note", "value": null, "literal": true}
  set (value equal to the current default): {"field": "demo_note", "value": "policy text"}
  clear: {"field": "demo_note", "value": null}
  reset: {"field": "demo_note", "absence_mode": "reset"}

  - The call is executed exactly once. There is no confirmation step, no retry and
    no repair of a rejected payload.
  - An entry that combines parameters the schema does not allow together is
    rejected and no part of the call is applied.
\end{verbatim}}

\noindent The \arm{Value+Prose(no examples)} instance omits the \code{EXAMPLES}
block. The convention-following call, which is also the aligned reference call,
touches one field:

{\footnotesize\begin{verbatim}
{"name": "update_record", "arguments": {"record_id": "DP-5202",
  "updates": [{"field": "reply_to_address", "value": "press@studio.example"}]}}
\end{verbatim}}

\noindent Under the counter-conventional contract that call resets the three
unmentioned fields. The displaced state has \code{signature\_line} inherited
(``Sent from the studio workspace'' at \code{dir-2031.4}) instead of absent, and
\code{label\_tags} inherited at \code{dir-2031.4} with effective value
\code{["directory", "studio"]} instead of at \code{dir-2030.9} with
\code{["directory"]}. \code{share\_photo} is unchanged only because it already
inherited the current revision. The counter-conventional reference call names
every field that must stay unchanged:

{\footnotesize\begin{verbatim}
{"name": "update_record", "arguments": {"record_id": "DP-5202",
  "updates": [{"field": "signature_line"},
              {"field": "reply_to_address", "value": "press@studio.example"},
              {"field": "label_tags"}, {"field": "share_photo"}]}}
\end{verbatim}}

\noindent The pair from the introduction is family
\code{T060-empty-fwd-contacts}: the request asks for the signature line to be
``present and holding the empty value \code{""}''. Under the aligned contract
(rule 4: an empty value is stored as the pinned value) the reference call's
entry for that field is \code{\{"field":"signature\_line","value":""\}}. Under
the counter contract (rule 4: ``this service treats an empty payload as an
erasure request'') that entry leaves the field absent, and the reference call
uses \code{\{"field":"signature\_line","literal":true,"value":""\}}. The
counter instance's examples block changes accordingly: \code{set (empty value)}
carries \code{"literal": true} and \code{clear} becomes
\code{\{"field": "demo\_note", "value": ""\}}. In both instances \code{keep}
uses \code{"absence\_mode": "keep"}.

\section{The five interfaces}
\label{app:interfaces}

The two overloaded interfaces share one entry schema. \code{field}, an
enumeration of the record's editable fields, is required. \code{value} has no
type in the schema; its type is given in the state block. \code{absence\_mode}
is an enumeration of \code{keep}, \code{clear} and \code{reset}, \code{literal}
is a boolean, and no additional properties are allowed. \arm{FlatDelta}
replaces the two control fields, \code{absence\_mode} and \code{literal}, by a
required \code{op} enumeration:

{\footnotesize\begin{verbatim}
{"type": "object", "additionalProperties": false, "required": ["field", "op"],
 "properties": {"field": {"type": "string", "enum": [...]},
                "op": {"type": "string", "enum": ["keep", "set", "clear", "reset"]},
                "value": {"description": "value payload ..."}}}
\end{verbatim}}

\noindent \arm{TypedDelta} nests the operation in a \code{delta} object whose
schema is a \code{oneOf} over four constructors, each with
\code{additionalProperties: false}, so that only \code{set} admits a value:

{\footnotesize\begin{verbatim}
{"type": "object", "additionalProperties": false, "required": ["field", "delta"],
 "properties": {"field": {...},
   "delta": {"oneOf": [
     {"required": ["op"],          "properties": {"op": {"const": "keep"}}},
     {"required": ["op", "value"], "properties": {"op": {"const": "set"}, "value": {...}}},
     {"required": ["op"],          "properties": {"op": {"const": "clear"}}},
     {"required": ["op"],          "properties": {"op": {"const": "reset"}}}]}}}
\end{verbatim}}

\noindent Both structural interfaces carry the same operation definitions, the
same truth table and the same seven worked examples as \arm{Value+Prose},
written in their own entry format. A note follows: rules 2--10 describe the
legacy compact entry, which this interface does not accept, and rule 1, the
omission rule, applies to every editable field either way. An entry that
combines \code{op} $\neq$ \code{set} with a value is rejected under
\arm{FlatDelta} (\code{op\_value\_conflict}) and cannot be written under
\arm{TypedDelta}.
Description lengths are audited per pair to be within $25\%$ across the three
documented interfaces. The measured means are $3021$, $3204$ and $3381$ characters
for \arm{Value+Prose}, \arm{FlatDelta} and \arm{TypedDelta}, within $12\%$ of
each other.

\section{Protocol details}
\label{app:protocol}

\paragraph{Models and transport.} Table~\ref{tab:models} maps every tier name
used in the generated tables to the model it ran. A \emph{cell} is one prompt
sent to one model. Every cell goes through one fixed transport, an agent
harness in single-prompt mode with tools, sessions, skills, rules, extensions
and retries disabled. It runs at temperature $0$ with thinking off, $1024$
output tokens and the shared system prompt of Appendix~\ref{app:pair}. No cell
uses a provider-native tool-calling path. The tool exists only as
documentation in the user message, and the call is a JSON object in the
assistant text. The harness prepends a fixed preamble (agent identity, working
directory, date) that cannot be disabled. The working directory is a fixed
path, so the preamble is the same for every cell. Each transport subprocess
runs with a closed \code{stdin} and with the parent's agent-session
environment variables removed. Concurrency is capped at two, and models run
one after another. A transport failure, meaning no assistant message at all,
is not scored, and the cell stays open to be sent again. A cell counts as done
once an assistant message exists, and that message is scored without changes.
The local models are served with speculative decoding disabled. The $27$B
model runs in NVFP4 and does not give identical outputs across runs, which is
why every held-out claim rests on 120 families rather than repeated trials.

\begin{table}[h]
\centering
\caption{Tier names used in the generated tables, and the models behind them.
Each held-out study has its own frozen model registry, so a later study that
adds a model never changes an earlier one.}
\label{tab:models}
\scriptsize
\begin{tabular}{l l l l}
\toprule
Tier & Model identifier & Study & Kind \\
\midrule
\code{frontier} & \code{claude-sonnet-4-5-20250929} & confirmatory, pilot & API, dated snapshot \\
\code{frontier\_strong} & \code{claude-opus-4-5} & confirmatory & API \\
\code{frontier\_alt} & \code{gpt-5.6-terra} & confirmatory & API, second vendor \\
\code{open\_strong} & Qwen3.8-27B (NVFP4, local) & confirmatory, pilot & open weights, pinned revision \\
\code{open\_small} & Qwen3-14B (local) & confirmatory, decision rule 7 & open weights, pinned revision \\
\code{frontier\_newest} & \code{claude-opus-5} & newer-model study & API \\
\code{frontier\_mid\_newest} & \code{claude-sonnet-5} & newer-model study & API \\
\code{frontier\_alt\_newest} & \code{gpt-daybreak-blue-latest} & newer-model study, second vendor & API \\
\bottomrule
\end{tabular}
\end{table}

\paragraph{Extraction and error taxonomy.} Three fixed strategies are tried in
order: the first fenced code block, the whole stripped response, and the
substring from the first \code{\{} to its matching \code{\}}. The candidate is
passed to a standard JSON parser unmodified. There is no trailing-comma
removal, quote fixing, key renaming, type coercion, default insertion or
retry, and a non-object top level is rejected. Outcomes are assigned with
fixed precedence.
\begin{itemize}\setlength{\itemsep}{0pt}
\item \code{correct}.
\item \code{parse\_failure}: no object extracted.
\item \code{malformed\_envelope}: wrong tool name, missing or non-object
  arguments, extra keys, missing record id or updates.
\item \code{abstention}.
\item \code{invalid\_call}: wrong record, duplicate field, unknown or
  read-only field, unknown entry key, \code{absence\_mode} combined with a
  value or with \code{literal}, \code{literal} without a value, wrong value
  type, and the structural-interface rejections of Appendix~\ref{app:interfaces}.
\item For calls that execute: \code{displaced\_convention} (the state the
  convention-following call reaches), \code{wrong\_operation},
  \code{wrong\_set\_value}, \code{collateral\_mutation},
  \code{latent\_only\_mismatch} (visibly correct, latently wrong) and
  \code{other\_wrong\_state}.
\end{itemize}
Every non-correct outcome counts as incorrect. In the $1200$ trials under
the terse description in the confirmatory study, the models abstained once
in each condition. One cell of the Anthropic newer-model study returned an
empty assistant message on six attempts. It was
scored as a parse failure; the gate figures are the same whether it is counted
or not.
In the confirmatory study, malformed output on counter-conventional instances
is at most $0.8\%$ for any model and interface, and neither \arm{Value+Prose} interface
produces an invalid call on a commercial model. The largest invalid rate,
$47.5\%$, belongs to the small local model under the terse description.

\paragraph{Statistics.} The unit of analysis is the task family. Every
interval is a paired bootstrap that resamples whole families ($10{,}000$
draws, fixed seed, percentile $[2.5, 97.5]$), so the two conditions and all
interfaces of a family move together. Paired comparisons also report an
exact McNemar test keyed on family and model. The convention-following error
rate (CFE) is the fraction of incorrect counter-conventional
\arm{ValueOnly} outputs whose executed state equals the displaced state, with
a Clopper--Pearson interval. Non-inferiority on aligned instances uses a $3$-point
margin on the lower confidence bound. Each pre-registration declares its own
Holm family: Q1--Q4 for the confirmatory study, G1--G2 and H1--H2 for the two
newer-model studies. Everything else is reported unadjusted and labelled as
such.

\paragraph{The four pre-registrations.} All four were written on the same day
and committed before their cells were sent.
\begin{itemize}\setlength{\itemsep}{2pt}
\item \emph{Pre-registration 1} (pilot: 60 development families,
  \arm{Value+Prose} versus \arm{FlatDelta} and \arm{TypedDelta} on two models,
  $720$ outputs) set an $8$-point gate for the structural interfaces and a $3$-point
  gate for a type-level claim. \arm{Value+Prose} scored $99.2\%$ aligned and
  $98.3\%$ counter, the structural interfaces could not gain, and the
  pre-registered decision was \textbf{STOP}. A development-only, ungated
  diagnostic of \arm{ValueOnly} on the same families then showed a large
  penalty concentrated in the \code{omission} family (per-family and per-model
  figures are in the released negative report).
\item \emph{Pre-registration 2} (confirmatory) declared the five models, the
  three interfaces, and four contrasts on the 120 held-out families.
  Q1 is \arm{ValueOnly} aligned minus counter, gated at $10$ points with a
  positive lower bound. Q2 is \arm{Value+Prose} minus \arm{ValueOnly} on
  counter-conventional instances, gated at $20$ points. Q3 is the same with the no-examples
  interface, reported and never gated. Q4 is Q1 restricted to \code{omission},
  gated at $30$ points. Its decision rules are as follows. A Q1 failure
  withdraws the paper (rule 3). Q3 within $5$ points of Q2 attributes the fix
  to the truth table, and Q3 within $10$ points of zero attributes it to
  example leakage (rules 4 and 5). Any model whose \arm{Value+Prose} counter
  accuracy is below $90\%$ must be run on the structural interfaces at the
  original $8$-point gate with the $3$-point aligned non-inferiority margin
  (rule 7). Section 7 of that document names the
  single number that could end the study: held-out counter-conventional
  \arm{ValueOnly} accuracy within $10$ points of aligned.
\item \emph{Pre-registration 3} declared \code{claude-opus-5} and
  \code{claude-sonnet-5} with the Q1 and Q2 thresholds (G1, G2), a rule that
  a smaller penalty on the newer model must be reported beside the headline
  with its numbers (G3), and an explicit prohibition on describing two pairs
  as a generational trend. It discloses an eight-cell development probe run
  before the document was written.
\item \emph{Pre-registration 4} declared \code{gpt-daybreak-blue-latest}
  with the same thresholds (H1, H2), the same reporting and trend rules, a rule
  deleting the single-vendor limitation if H1 passes, and a rule that a
  provider outage leaves the study incomplete rather than counting as a
  result.
\end{itemize}

\paragraph{Audit checks and release.} The release bundle contains the compiler,
the frozen manifests, the $12{,}167$ deterministic audit checks and outcome
records for all $7200$ scored traces. We prepared it for release at
\url{https://github.com/NaiveNeuron/LeaveItOutBench}.
The checks ($8{,}111$ held out, $4{,}056$ development) are computed by a
script, not a model, and all pass. Per pair they verify the following.
\begin{itemize}\setlength{\itemsep}{0pt}
\item Identical intent, initial state, defaults, target and per-interface schemas.
\item Exactly one differing contract rule, equal to the designated rule, and
  totality and surjectivity of both contracts.
\item No operation word in any field or flag name and no operation
  constructor in the request.
\item Every interface's reference call reaching the target, and the
  convention-following call correct under the aligned contract and wrong,
  with a latent difference, under the counter contract.
\item \op{Set}(default) producing an override and \op{Reset} an inheritance
  at the current revision. \op{Set}(empty) and \op{Set}(null) producing
  present fields and \op{Clear} an absent one.
\item Documented interfaces stating all four operation words, all ten rules and
  the same seven examples with no record field leaking into them, and
  description lengths within $25\%$.
\end{itemize}

\paragraph{The unexecuted transfer suite.} We adapted ToolSandbox's stateful
multi-turn setting into 48 declarative scenarios over a world-state store. A
deterministic scripted user replaces its model-simulated user, so no model
appears on the scoring side, and we verified the harness end to end with a
stub transport that replays reference calls. The 288 planned episodes were
never sent: the runner does not start while the pilot report records a
\textbf{STOP} decision, and no later pre-registration re-authorised the
suite. The prepared release bundle includes the transfer suite.

\section{Full result tables}
\label{app:tables}

All tables are generated from the frozen summaries. They name models by tier
name (Table~\ref{tab:models}), write ``Counter'' for counter-conventional and,
in Table~\ref{tab:h5}, ``Arm'' for interface. CIP is aligned minus
counter-conventional accuracy in percentage points.

\begin{table}[h]
\centering
\caption{Confirmatory study: exact accuracy per model and interface on the
120 held-out families.}
\scriptsize
\begin{tabular}{llrrrr}
\toprule
Model & Interface & Aligned & Counter & CIP & $n$ \\
\midrule
frontier & ValueOnly & 88.3 & 60.0 & 28.3 & 240 \\
frontier & Value+Prose & 100.0 & 97.5 & 2.5 & 240 \\
frontier & Value+Prose(no examples) & 100.0 & 97.5 & 2.5 & 240 \\
frontier\_alt & ValueOnly & 93.3 & 62.5 & 30.8 & 240 \\
frontier\_alt & Value+Prose & 100.0 & 100.0 & 0.0 & 240 \\
frontier\_alt & Value+Prose(no examples) & 100.0 & 100.0 & 0.0 & 240 \\
frontier\_strong & ValueOnly & 90.0 & 51.7 & 38.3 & 240 \\
frontier\_strong & Value+Prose & 100.0 & 100.0 & 0.0 & 240 \\
frontier\_strong & Value+Prose(no examples) & 100.0 & 99.2 & 0.8 & 240 \\
open\_small & ValueOnly & 43.3 & 28.3 & 15.0 & 240 \\
open\_small & Value+Prose & 87.5 & 70.0 & 17.5 & 240 \\
open\_small & Value+Prose(no examples) & 87.5 & 55.0 & 32.5 & 240 \\
open\_strong & ValueOnly & 65.8 & 45.0 & 20.8 & 240 \\
open\_strong & Value+Prose & 100.0 & 98.3 & 1.7 & 240 \\
open\_strong & Value+Prose(no examples) & 99.2 & 95.0 & 4.2 & 240 \\
\midrule
\textit{pooled} & ValueOnly & 76.2 & 49.5 & 26.7 & -- \\
\textit{pooled} & Value+Prose & 97.5 & 93.2 & 4.3 & -- \\
\textit{pooled} & Value+Prose(no examples) & 97.3 & 89.3 & 8.0 & -- \\
\bottomrule
\end{tabular}

\end{table}

\begin{table}[h]
\centering
\caption{Confirmatory study, \arm{ValueOnly} by collision family, pooled over
the five models. $n$ is the number of outputs over both conditions.}
\scriptsize
\begin{tabular}{lrrrr}
\toprule
Collision family & Aligned & Counter & CIP & $n$ \\
\midrule
\texttt{omission} & 76.9 & 3.8 & 73.1 & 320 \\
\texttt{null} & 75.7 & 70.0 & 5.7 & 280 \\
\texttt{empty} & 72.1 & 62.1 & 10.0 & 280 \\
\texttt{default} & 79.4 & 66.2 & 13.1 & 320 \\
\bottomrule
\end{tabular}

\end{table}

\begin{table}[h]
\centering
\caption{Counter-conventional \code{omission} families: what the models wrote
for each field that had to stay unchanged (a keep-target slot). Slots that are
neither an explicit \code{keep} nor omitted hold some other entry.}
\scriptsize
\begin{tabular}{lrrr}
\toprule
Interface & Keep-target slots & Explicit \texttt{keep} & Omitted \\
\midrule
ValueOnly & 400 & 26 & 306 \\
Value+Prose & 400 & 327 & 60 \\
Value+Prose(no examples) & 400 & 331 & 61 \\
\bottomrule
\end{tabular}

\end{table}

\begin{table}[h]
\centering
\caption{Anthropic newer-model study (pre-registration 3): \arm{ValueOnly}
accuracy of each pre-registered (``declared'') model and its newer sibling.
Intervals are 95\% bootstrap intervals on the CIP that resample whole task
families.}
\label{tab:gen1}
\scriptsize
\begin{tabular}{llrrrl}
\toprule
Family & Tier & Aligned & Counter & CIP & 95\% CI \\
\midrule
declared & frontier\_strong & 90.0 & 51.7 & 38.3 & [30.0, 47.5] \\
newer & \textbf{frontier\_newest} & 78.3 & 57.5 & 20.8 & [13.3, 29.2] \\
\midrule
declared & frontier & 88.3 & 60.0 & 28.3 & [20.0, 36.7] \\
newer & \textbf{frontier\_mid\_newest} & 90.8 & 67.5 & 23.3 & [14.2, 32.5] \\
\bottomrule
\end{tabular}

\end{table}

\begin{table}[h]
\centering
\caption{Second-vendor newer-model study (pre-registration 4), laid out as
Table~\ref{tab:gen1}.}
\scriptsize
\begin{tabular}{llrrrl}
\toprule
Family & Tier & Aligned & Counter & CIP & 95\% CI \\
\midrule
declared & frontier\_alt & 93.3 & 62.5 & 30.8 & [21.7, 40.0] \\
newer & \textbf{frontier\_alt\_newest} & 90.8 & 58.3 & 32.5 & [22.5, 42.5] \\
\bottomrule
\end{tabular}

\end{table}

\begin{table}[h]
\centering
\caption{Decision rule 7 on the one eligible model: counter-conventional
accuracy of each structural interface and its gain over \arm{Value+Prose},
against the pre-registered $8$-point gate.}
\label{tab:h5}
\scriptsize
\begin{tabular}{llrrrl}
\toprule
Model & Arm & Counter & vs Prose & 95\% CI & 8\,pp \\
\midrule
open\_small & FlatDelta & 75.8 & 5.8 & [2.5, 10.0] & no \\
open\_small & TypedDelta & 76.7 & 6.7 & [1.7, 11.7] & no \\
\bottomrule
\end{tabular}

\end{table}

\end{document}